\documentclass[conference]{IEEEtran}
\IEEEoverridecommandlockouts
\usepackage{cite}
\usepackage{amsmath,amssymb,amsfonts}
\usepackage{algorithmic}
\usepackage{graphicx}
\usepackage{textcomp}
\usepackage{xcolor}
\usepackage{hyperref}
\usepackage{booktabs}
\usepackage{acro}

\DeclareAcronym{JSSP}{
  short = JSSP,
  long  = Job Shop Scheduling Problem
}
\DeclareAcronym{PORL}{
  short = PORL,
  long  = Pretrained Offline Reinforcement Learning
}
\DeclareAcronym{DQN}{
  short = DQN,
  long  = Deep Q-Network
}
\DeclareAcronym{RL}{
  short = RL,
  long  = Reinforcement Learning
}
\DeclareAcronym{CQL}{
  short = CQL,
  long  = Conservative Q Learning
}
\DeclareAcronym{GNN}{
  short = GNN,
  long  = Graph Neural Network
}
\DeclareAcronym{KL}{
  short = KL,
  long  = Kullback-Leibler
}
\DeclareAcronym{GIN}{
  short = GIN,
  long  = Graph Isomorphism Network
}
\DeclareAcronym{FDDMWR}{
  short = FDD/MWR,
  long = Flow Due Date / Most Work Remaining
}

\def\BibTeX{{\rm B\kern-.05em{\sc i\kern-.025em b}\kern-.08em
    T\kern-.1667em\lower.7ex\hbox{E}\kern-.125emX}}
\begin{document}

\title{PORL: Pretrained Offline Reinforcement Learning for the Job Shop Scheduling Problem\\
\thanks{Supported by the Chips Joint Undertaking and its members, including top-up funding by National Authorities, within the Cynergy4MIE project (Grant Agreement No. 101140226).}
}

\author{\IEEEauthorblockN{1\textsuperscript{st} Mateo Toro Diz}
\IEEEauthorblockA{\textit{Department of Industrial Engineering} \\
\textit{Rosenheim University}\\
\textit{of Applied Sciences} \\
Rosenheim, Germany \\
mateo.toro-diz@th-rosenheim.de}
\and
\IEEEauthorblockN{2\textsuperscript{nd} Jonathan Hoss}
\IEEEauthorblockA{\textit{Department of Industrial Engineering} \\
\textit{Rosenheim University}\\
\textit{of Applied Sciences} \\
Rosenheim, Germany\\
jonathan.hoss@th-rosenheim.de}
\and
\IEEEauthorblockN{3\textsuperscript{rd} Noah Klarmann}
\IEEEauthorblockA{\textit{Department of Industrial Engineering} \\
\textit{Rosenheim University}\\
\textit{of Applied Sciences} \\
Rosenheim, Germany \\
noah.klarmann@th-rosenheim.de}
}

\maketitle

\begin{abstract}
The Job Shop Scheduling Problem (JSSP) is a fundamental combinatorial optimization problem in industrial optimization. This work introduces Pretrained Offline Reinforcement Learning (PORL), a hybrid approach that combines simulation-based online pretraining with offline fine-tuning on production-specific data. 

Reinforcement learning through online interaction enables exploration of general scheduling strategies, but typically relies on simulation environments and may suffer from a simulation-to-reality gap. In contrast, offline RL avoids direct interaction with the environment by learning from historical data, but its performance is strongly influenced by dataset quality and coverage. PORL combines the strengths of both paradigms by first learning a general scheduling policy through online interaction and subsequently adapting it offline to a target distribution. A KL-divergence-based policy constraint is introduced to limit deviations from the pretrained policy during fine-tuning. 

The approach is evaluated on JSSP instances with distribution shift and datasets generated from heuristic, noisy-expert, and random behavioral policies. The results show that PORL consistently achieves lower optimality gaps than standalone offline RL and the considered general scheduling baselines. Furthermore, its advantage over standalone offline RL increases as dataset quality decreases, indicating reduced sensitivity to the quality and coverage of the available offline data. The results suggest that offline adaptation of pretrained policies is a promising approach for industrial scheduling environments where direct online exploration is impractical.

\end{abstract}


\section{Introduction}
The \ac{JSSP} is a classical combinatorial optimization problem in the field of operations research and management science. Solving the \ac{JSSP} involves scheduling jobs and their constituent operations on available machines to minimize the total production time, commonly referred to as the makespan. \ac{JSSP} is highly relevant to industrial production systems, as it serves as an idealized representation of the scheduling challenges encountered on manufacturing shop floors.

The classical \ac{JSSP} defines the scheduling problem as a static instance, where the set of jobs does not change over time. When this assumption holds,  methods such as Branch \& Bound \cite{brucker_branch_nodate,artigues_branch_2008,nababan_branch_2008} and Constraint Programming \cite{da_col_industrial-size_2022} can produce optimal or near-optimal schedules for many benchmark instances, although at a high computational cost due to the NP-hardness of the problem. However, real shop floors tend to operate in dynamic environments, where heuristic approaches are often preferred, ranging from simple Priority Dispatching Rules \cite{habib_zahmani_multiple_2015} to more sophisticated meta-heuristics \cite{hajariwala_review_2025}.

Despite the success of classical approaches, their reliance on handcrafted dispatching rules or computationally intensive instance-specific optimization limits their ability to adapt to changing production environments and generalize across diverse scheduling scenarios. This has motivated the investigation of learning-based approaches, particularly Reinforcement Learning (RL), which learns scheduling policies directly from experience. More recently, \ac{RL} has been successfully applied to \ac{JSSP}, achieving competitive scheduling performance while learning policies that can generalize across multiple problem instances \cite{zhang_learning_2020,park_learning_2021,maharjan_reinforcement_2026}. However, online \ac{RL} approaches require a simulation environment for training, introducing a simulation-to-reality gap.

Offline \ac{RL} aims to reduce this dependency on simulation by learning directly from previously collected trajectories, eliminating the need for environment interaction during training. However, its performance remains highly dependent on the quality and coverage of the available dataset \cite{levine_offline_2020, schweighofer_dataset_2022, remmerden_generalizing_2025}.

This paper introduces Pretrained Offline Reinforcement Learning (PORL), a hybrid framework that combines the exploration capabilities of online \ac{RL} with the adaptability of offline \ac{RL}. \ac{PORL} first trains a scheduling policy through online interaction with a simulated environment, enabling the agent to acquire general scheduling strategies. The resulting policy is subsequently refined using offline \ac{RL} on production-specific datasets, allowing adaptation to the target shop floor without requiring online exploration. By initializing offline learning from a policy that has already explored a large portion of the scheduling state space, \ac{PORL} reduces the dependence on dataset coverage and mitigates one of the key limitations of conventional offline RL.

Unlike conventional hybrid \ac{RL} approaches, which employ offline pretraining followed by online fine-tuning, \ac{PORL} adopts the reverse training order. This design is motivated by the high cost and operational risk of exploratory scheduling decisions in real-world industrial systems, where learning directly through interaction with the deployed production environment is typically infeasible.

Our work offers the following contributions:
\begin{itemize}
\item We introduce \ac{PORL}, a novel approach combining simulation-based online pretraining with log-based offline fine-tuning.
\item We propose a policy-constrained offline fine-tuning objective that adapts pretrained Q-networks using offline data while controlling divergence from the pretrained policy.
\item We show that \ac{PORL} achieves lower optimality gaps than both dispatching-rule baselines and standalone offline \ac{RL} approaches under significant distributional shift.
\item We demonstrate that \ac{PORL} maintains strong performance even when trained on low-return datasets, outperforming standalone offline \ac{RL} approaches.
\end{itemize}

\section{Related work}
\subsection{\ac{RL} for \ac{JSSP}}
Recent work formulates the \ac{JSSP} as a sequential decision-making task, enabling the application of \ac{RL} methods \cite{zhang_learning_2020,park_learning_2021,maharjan_reinforcement_2026}. Most approaches employ graph-based state representations and a \ac{GNN} to capture precedence and machine constraints, achieving strong performance and generalization on benchmark instances \cite{zhang_learning_2020,park_learning_2021,munikoti_challenges_2022, hoss_scalable_2026}. Among them, Deep Q-Networks (DQN) combine temporal-difference learning with experience replay and target networks to learn scheduling policies. However, these methods are predominantly trained through online interaction with simulation environments, making them dependent on simulator fidelity and limiting their ability to leverage historical production data.

\subsection{Offline \ac{RL} for scheduling}
Offline \ac{RL} learns policies from fixed datasets without environment interaction, making it attractive for industrial scheduling applications where exploration is costly. A central challenge is distributional shift, which can lead to unreliable value estimates for actions not represented in the dataset \cite{levine_offline_2020}. \ac{CQL} addresses this issue through value regularization and has become a widely used offline \ac{RL} algorithm \cite{kumar_conservative_2020}.

The application of offline \ac{RL} to \ac{JSSP} remains relatively recent. Remmerden et al. \cite{remmerden_offline_2025,remmerden_generalizing_2025} demonstrate that \ac{CQL}-based scheduling policies with action masking can achieve competitive performance when trained on datasets generated from optimal, heuristic, and noisy trajectories. Echeverria et al. \cite{echeverria_offline_2025} further explore policy-regularized offline \ac{RL} for scheduling. Despite these advances, offline \ac{RL} remains fundamentally constrained by dataset quality and coverage.

\subsection{Research gap}
Current \ac{RL} approaches to \ac{JSSP} exhibit a fundamental trade-off. Online \ac{RL} enables exploration and discovery of improved scheduling strategies but relies on simulation environments that may not accurately reflect real production systems. Offline \ac{RL} leverages historical production data but is constrained by the support of the available dataset.

Hybrid \ac{RL} approaches in other domains typically combine offline pretraining with online fine-tuning \cite{nair_awac_2021,yao_control_2023,chen_offline--online_2025}. However, this paradigm assumes that online exploration in the target environment is feasible, which is often unrealistic in industrial scheduling due to the cost of suboptimal decisions. To the best of our knowledge, the reverse paradigm—online pretraining in simulation followed by offline adaptation to production data—has not been systematically explored for \ac{JSSP}. This gap motivates the \ac{PORL} framework proposed in this work.

\section{Algorithm}
\begin{figure}[t]
\centering
\includegraphics[width=\columnwidth]{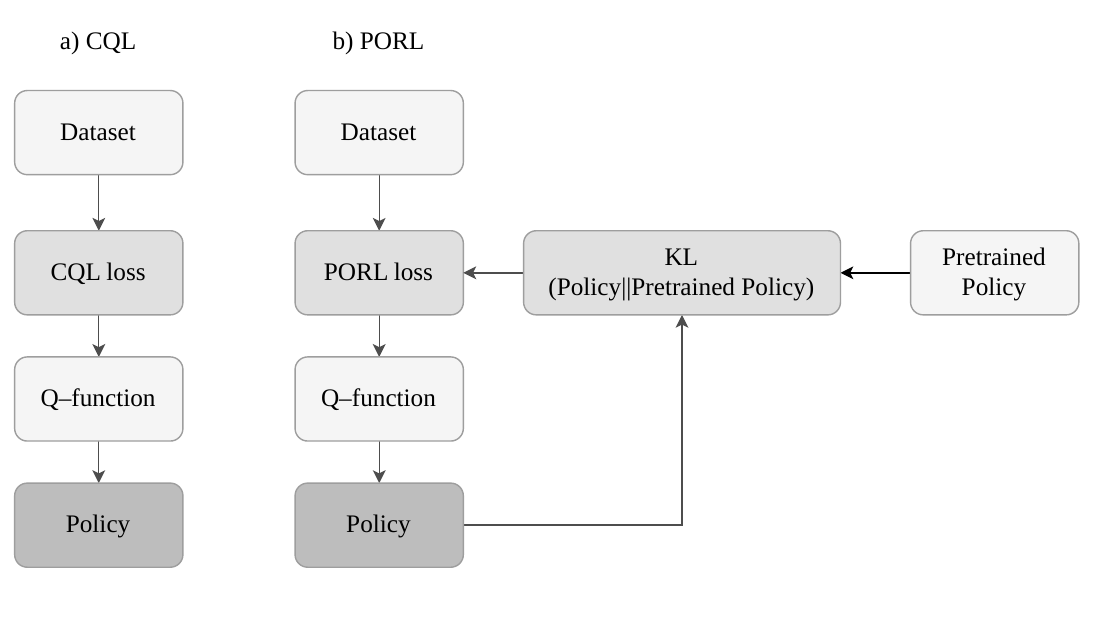}
\caption{Training pipeline of (a) the offline \ac{RL} baseline and (b) the proposed \ac{PORL} approach.}
\label{fig:algopipeline}
\end{figure}

Fig. \ref{fig:algopipeline} illustrates the overall training pipeline of the proposed method and the offline baseline.

\subsection{Model Architecture}
\label{sec:architecture}

We use the graph representation and \ac{GNN} architecture of Hoss, Link, and Klarmann \cite{hoss_scalable_2026}, where each \ac{JSSP} state $s_t$ is modeled as a heterogeneous graph $G_t = (V, E)$ consisting of operation and machine nodes:

\begin{equation}
\begin{aligned}
V = V_{\mathrm{ops}} \cup V_{\mathrm{mch}}.
\end{aligned}
\end{equation}

The graph contains two edge types. Precedence edges connect consecutive operations within a job, with at most two such edges per operation. Assignment edges link each operation to its corresponding machine, with at most one assignment edge per operation. This sparse representation results in a number of edges that scales linearly with the number of operations, avoiding the quadratic edge growth of dense disjunctive graph representations.

The proposed \ac{GNN} operates on the graph using a feature-based homogenization scheme. Node-type information, distinguishing operations from machines, is encoded via one-hot embeddings. This allows both node types to be processed in a shared feature space while preserving their structural distinctions.

The resulting homogeneous graph is processed using a $K$-layer \ac{GIN}\cite{xu_how_2019}, where node embeddings are updated via multilayer perceptrons. Stacking layers enables aggregation over $K$-hop neighborhoods, allowing the model to capture both local scheduling constraints and global job-machine interactions.

For all experiments, we use a fixed architecture with hyperparameters listed in Table~\ref{tab:networkarchitecture}.

\begin{table}[tbp]
\centering
\caption{GNN architecture hyperparameters.}
\label{tab:networkarchitecture}
\begin{tabular}{ll}
\toprule
\textbf{Parameter} & \textbf{Value} \\
\midrule
Input feature dimension & 5 \\
Hidden dimension & 64 \\
Number of \ac{GIN} layers ($K$) & 3 \\
\bottomrule
\end{tabular}
\end{table}

\subsection{Pretrained Offline \ac{RL}}
\subsubsection{Positioning}
We propose a hybrid training approach that combines online pretraining with offline fine-tuning, and name it Pretrained Offline RL (PORL). The online phase enables the agent to learn a general scheduling strategy from diverse simulated instances, while the offline phase adapts this policy to the specific distribution of target instances derived from industrial data. The fine-tuning process is constrained to prevent large deviations from the pretrained policy, thereby preserving previously learned high-quality behaviors while allowing targeted specialization.

From an industrial perspective, \ac{PORL} is intended to specialize a well-trained general policy to a specific production setting using limited offline data. This work evaluates the feasibility of this adaptation principle; the online and offline training budgets are therefore kept fixed.

\subsubsection{Initialization and setup}
The \ac{PORL} training process is initialized from a pretrained online DQN agent. The parameters learned during the online phase are directly transferred to the hybrid agent by loading the network weights. This transfer requires an identical Q-network architecture across both training stages. The transferred agent is then fine-tuned offline using the policy-constrained objective described below.

\subsubsection{Policy Constraint Formulation}
\label{sec:porlformulation}

During offline fine-tuning, the objective is to adapt the pretrained Q-function to the distribution of the target instances while preventing large deviations from the original policy.

To achieve this, we introduce a policy-level regularization term that constrains the updated policy to remain close to the pretrained reference policy. This approach is conceptually inspired by policy-regularized offline \ac{RL} methods \cite{fujimoto_off-policy_2019}, which similarly enforce proximity between the learned policy and a reference distribution—typically the behavior policy \( \pi_\beta \)—to mitigate distributional shift and ensure stable learning. In contrast, our method uses the pretrained online policy as the reference, enabling controlled adaptation while preserving previously acquired knowledge.

Since DQN is a value-based method, the policy is not learned explicitly but is implicitly defined through the Q-values. To enable a comparison between policies, we derive a stochastic policy from the Q-function using a softmax transformation:
\begin{equation}
\begin{aligned}
\pi(a \mid s) =
\begin{cases}
\dfrac{\exp\!\left(\frac{Q(s,a)}{\tau}\right)}{\sum\limits_{a' \in \mathcal{A}(s)} \exp\!\left(\frac{Q(s,a')}{\tau}\right)}, & a \in \mathcal{A}(s), \\[1.2em]
0, & a \notin \mathcal{A}(s),
\end{cases}
\label{eq:policy-softmax}
\end{aligned}
\end{equation}
where \( Q(s, a) \) denotes the predicted action-value, \( \tau \) is a temperature parameter controlling the smoothness of the distribution, and \( \mathcal{A}(s) \) is the set of valid actions defined by the action mask. The reference policy $\pi_{\mathrm{ref}}$ is obtained by applying the same softmax transformation to the pretrained Q-function, and remains fixed throughout the fine-tuning process.

The deviation between the fine-tuned policy \( \pi_{\mathrm{new}} \) and the pretrained reference policy \( \pi_{\mathrm{ref}} \) is then quantified using the Kullback--Leibler (KL) divergence. We employ the forward KL divergence $D_{\mathrm{KL}}(\pi_{\mathrm{new}} \parallel \pi_{\mathrm{ref}})$, which penalizes assigning probability mass to actions that are unlikely under the reference policy, while still allowing the updated policy to refine and reweight action preferences, thereby encouraging improvement on high-value actions while discouraging the selection of low-value ones. The resulting regularization term is defined as:
\begin{equation}
\begin{aligned}
\mathcal{L}_{\mathrm{KL}} =
\lambda_{\mathrm{KL}} \frac{1}{B}
\sum_{i=1}^{B}
\sum_{\!a \in \mathcal{A}(s_i)}
\pi_{\mathrm{new}}(a|s_i)
\Big[
\log(\pi_{\mathrm{new}}(a|s_i)+\varepsilon) \\
- \log(\pi_{\mathrm{ref}}(a|s_i)+\varepsilon)
\Big]
\end{aligned}
\label{eq:kl-loss}
\end{equation}
where \( \lambda_{\mathrm{KL}} \) controls the strength of the constraint, \( B \) denotes the batch size, and \( \varepsilon \) is a small constant introduced for numerical stability.

In addition, we retain the \ac{CQL} loss, since out-of-distribution actions remain undesirable in the proposed online-to-offline training pipeline.

The overall training objective combines the DQN loss, CQL regularization, and the KL policy constraint:
\begin{equation}
\mathcal{L} = \mathcal{L}_{\text{DQN}} + \mathcal{L}_{\text{CQL}} + \mathcal{L}_{\mathrm{KL}}.
\label{eq:hybrid-loss}
\end{equation}

where $\mathcal{L}_{\text{DQN}}$ denotes the temporal-difference loss defined in Equation \ref{eq:dqn-loss}, while the CQL loss is defined in Equation \ref{eq:cql-objective}.
\begin{equation}
\begin{aligned}
\mathcal{L}_{\text{DQN}} &=
\mathbb{E}_{(s,a,r,s') \sim \mathcal{D}}
\big[(Q(s,a)-y)^2\big], \\
y &= r + \gamma \max_{a'} Q(s',a')
\end{aligned}
\label{eq:dqn-loss}
\end{equation}

\begin{equation}
\mathcal{L}_{\text{CQL}}
=
\alpha\!\left(
\mathbb{E}_{s\sim\mathcal{D}}
\!\left[\log\!\sum_a e^{Q(s,a)}\right]
-
\mathbb{E}_{(s,a)\sim\mathcal{D}}[Q(s,a)]
\right)
\label{eq:cql-objective}
\end{equation}

This formulation limits deviations from the pretrained policy while still allowing targeted adaptation to the new data distribution, thereby enabling controlled specialization without catastrophic degradation of previously learned behavior.

\section{Dataset generation}
\label{sec:instances}
We denote the size of a \ac{JSSP} instance as $J \times M$, where $J$ is the number of jobs, and $M$ is the number of machines. 

Online pretraining is performed on standard, uniformly distributed $15 \times 15$ \ac{JSSP} instances.

To emulate adaptation to a specific industrial shop floor, offline fine-tuning uses custom instance distributions that introduce a significant distribution shift.

The generated shop floor templates define (i) machine-specific processing time distributions and (ii) job precedence constraints. Processing times are sampled from machine-dependent Gaussian or uniform distributions, while 50\% of the jobs are assigned partial or full precedence constraints. Partial constraints enforce only selected machine ordering relations, whereas full constraints specify a complete machine sequence for a job. The remaining jobs follow the standard unconstrained \ac{JSSP} formulation. The instance size is also shifted with respect to online pretraining, with instance sizes of $10 \times 10$.

This setup produces specialized instance distributions that differ substantially from the uniformly distributed instances commonly used during online pretraining, which were originally designed as generic optimization benchmarks. In contrast, the proposed specialized instances more closely resemble real-world shop floors, where specific product flows occur repeatedly, and machines exhibit heterogeneous characteristics. As a result, they introduce a realistic distribution shift that helps bridge the gap between simulation-based training and practical manufacturing environments.

\subsection{Behavioral policies}
\label{sec:behpolicies}
To simulate scheduling decisions observed in industrial environments, three different sets of behavioral policies are introduced, ranging from random actions to exact-solver solutions, replicating previous works in the literature \cite{remmerden_generalizing_2025}

The first dataset is generated using two $\epsilon$-greedy priority dispatching rules. The policies follow the corresponding heuristic while selecting a random valid action with probability $\epsilon$, introducing variability similar to that found in real production logs.

The second set of behavioral policies is based on Remmerden, Bukhsh, and Zhang's "Noisy-expert" policies \cite{remmerden_offline_2025}. This policy follows a schedule produced by an exact solver with probability $0.5$. When not, for each scheduled action, there is a $0.1$ probability that the agent will take a random valid action. When a previously scheduled random action prevents the agent from taking the scheduled optimal action, another random valid action is scheduled.

The third  behavioral policy is formed by random valid actions to simulate a worst-case scenario where the production logs contain low-quality schedules.

The hyperparameters are kept constant across all 3 datasets to test the robustness of the proposed \ac{PORL} algorithm to reduced-quality datasets.

\section{Agent training}

This section describes the training protocols used for the online \ac{RL}, standalone offline \ac{RL}, and \ac{PORL} agents. To enable a controlled comparison, all approaches use the same underlying network architecture, while their training data and optimization procedures differ according to the respective learning distributions, training budgets, and hyperparameter used for each approach. 

\subsection{Online \ac{RL} training}
The online \ac{RL} baseline is trained on a fixed set of 100 $15 \times 15$ uniformly distributed \ac{JSSP} instances, with operation durations ranging from 0 to 100. The agent learns with the DQN algorithm, based on the architecture from section \ref{sec:architecture} with the hyperparameters listed in \autoref{tab:online_dqn_learning_exp15}.

\begin{table}[tbp]
\centering
\caption{Online DQN learning hyperparameters.}
\label{tab:online_dqn_learning_exp15}
\begin{tabular}{ll}
\toprule
\textbf{Hyperparameter} & \textbf{Value} \\
\midrule
Learning rate & $1.5 \times 10^{-4}$ \\
Discount factor ($\gamma$) & 0.99 \\
TD loss function & smooth\_l1 \\
CQL regularization ($\alpha$) & 0.0 \\
Training timesteps & 250000 \\
Replay buffer size & 10000 \\
Learning starts after & 2000 \\
Batch size & 64 \\
Training frequency (env steps/update) & 4 \\
Target network update frequency & 500 \\
$\epsilon$ (start/end/decay) & 1.0/0.05/125000 \\
\bottomrule
\end{tabular}
\end{table}

\subsection{Offline and \ac{PORL}}
Both the standalone offline \ac{RL} baseline and the proposed \ac{PORL} agent are trained on the target-specific instances described in \autoref{sec:instances}. To approximate production logs, the offline datasets are generated using the stochastic behavioral policies introduced in \autoref{sec:behpolicies}. Each behavioral policy is rolled out on 50 instances sampled from the target distribution.

The standalone offline \ac{RL} baseline is trained from scratch using CQL-DQN. In contrast, \ac{PORL} initializes the Q-network with the weights obtained during online pretraining and subsequently fine-tunes the agent on the same offline dataset. This setup enables a direct comparison between learning exclusively from offline data and adapting a pretrained policy to the target distribution.

\subsubsection{Offline RL}
The offline \ac{RL} algorithm is trained with fixed hyperparameters, as shown in \autoref{tab:offline_cql_learning}.

\begin{table}[tbp]
\centering
\caption{Offline CQL-DQN learning hyperparameters.}
\label{tab:offline_cql_learning}
\begin{tabular}{ll}
\toprule
\textbf{Hyperparameter} & \textbf{Value} \\
\midrule
Learning rate & $1.0 \times 10^{-5}$ \\
Discount factor ($\gamma$) & 0.99 \\
Training timesteps & 100000 \\
TD loss function & smooth\_l1 \\
CQL regularization ($\alpha$) & 1.0 \\
Batch size & 64 \\
Target network update frequency & 500 \\
\bottomrule
\end{tabular}
\end{table}

\subsubsection{Pretrained Offline RL}
\label{sec:porltraining}
The \ac{PORL} agent is trained with the same dataset as the offline one, and is provided with the same amount of training steps, to enable a direct comparison between both approaches. The corresponding hyperparameters are shown in \autoref{tab:hybrid_cql_learning}.

\begin{table}[tbp]
\centering
\caption{Pretrained Offline agent learning hyperparameters.}
\label{tab:hybrid_cql_learning}
\begin{tabular}{ll}
\toprule
\textbf{Hyperparameter} & \textbf{Value} \\
\midrule
$\lambda_{\mathrm{KL}}$ & 0.5 \\
Discount factor ($\gamma$) & 0.99 \\
Training timesteps & 100000 \\
TD loss function & smooth\_l1 \\
Batch size & 64 \\
Target network update frequency & 500 \\
\bottomrule
\end{tabular}
\end{table}

To reduce the instability inherent to the multiobjective optimization characteristic of the presented constrained DQN algorithm, the values of both the learning rate and CQL $\alpha$ are changed with time. The specific ramping methods are displayed in \autoref{tab:hybrid_cql_ramping}.

\begin{table}[tbp]
\centering
\caption{PORL hyperparameters ramping.}
\label{tab:hybrid_cql_ramping}
\begin{tabular}{lcc}
\toprule
\textbf{Parameter} & $\alpha_{\mathrm{CQL}}$ & Learning rate \\
\midrule
Initial value & 0.0 & $5.0 \times 10^{-6}$ \\
Final value & 1.0 & $2.0 \times 10^{-6}$ \\
Ramp start & 0 & 40000 \\
Ramp end & 40000 & 80000 \\
\bottomrule
\end{tabular}
\end{table}

\section{Evaluation Protocol}
To account for variability due to stochastic training, each model is trained using five independent random seeds. The same seeds are used for corresponding offline and \ac{PORL} experiments to ensure consistency and comparability.

To test the generalization of the agent to new instances, the evaluation of the performance is done with a total of 100 newly generated instances, which are themselves created with 5 seeds, which also differ from the training ones. The evaluation instances are kept constant between experiments and approaches.

The performance of the trained agents is measured in terms of makespan and optimality gap. The makespan is calculated as the average makespan achieved across all evaluation instances, averaged between differently seeded experiments.

Additionally, to improve clarity, the performance is also measured in terms of the optimality gap. The optimality gap is defined as: 
\begin{equation}
\mathrm{Optimality\ Gap}\,(\%) = \frac{C - C^{*}}{C^{*}} \cdot 100
\label{eq:optimality-gap}
\end{equation}
where $C$ is the makespan achieved by the evaluated agent and $C^{*}$ is the optimal makespan obtained by the exact solver, with a time limit of 300 seconds.

\subsection{Benchmarking}
To validate the performance of both the offline and hybrid agents, there are several benchmarks against which the performance is to be compared.

By the use of the optimality gap metric, a comparison against the optimal solver is already implicit. Additionally, we also compare the results against the results achieved by:
\begin{itemize}
	\item Priority Dispatching Rules
	\item A state-of-the-art PPO agent trained on benchmark, uniformly distributed $20\times20$ \ac{JSSP} instances, following the work of Hoss, Link, and Klarmann \cite{hoss_scalable_2026}.
\end{itemize}

\section{Results}
The results in \autoref{tab:main_results} show that the proposed hybrid \ac{PORL} setup achieves lower optimality gaps when compared to both the standard offline \ac{RL} approach and commonly used general solutions to the JSSP, including the \ac{FDDMWR} priority dispatching rule and an online-trained \ac{RL} agent \cite{hoss_scalable_2026}). The results displayed in \autoref{tab:main_results} are averaged across all 5 training seeds, which remain the same for both offline and PORL.

The proposed \ac{PORL} pipeline achieves a relative reduction of $7.5\%$ in the optimality gap when compared to classical offline RL, and $28.7\%$ when compared to a state-of-the-art PPO agent trained on traditional, uniformly distributed \ac{JSSP} instances.

\begin{figure}[tbp]
\centering
\includegraphics[width=\columnwidth]{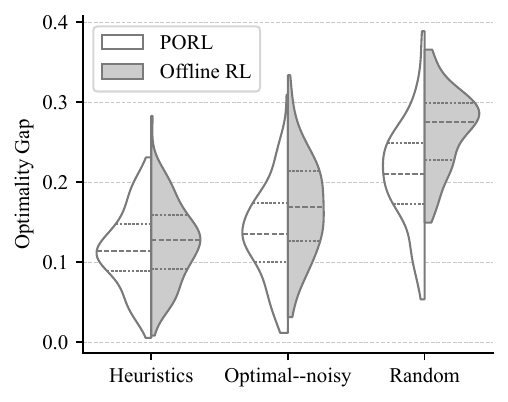}
\caption{PORL and Offline RL optimality gap comparison}
\label{fig:violinplot}
\end{figure}

The standard deviation of the optimality gap across seeds is substantially smaller than the observed performance difference between Offline \ac{RL} and \ac{PORL}, and at the same time, slightly smaller than the standard deviation of Offline RL. The distribution of the optimality gap for both \ac{PORL} and offline \ac{RL} can be seen in Fig. \ref{fig:violinplot}.

\begin{table}[tbp]
\centering
\caption{Makespan optimality gap (\%) for different scheduling approaches, evaluated on high distribution shift instances.}
\label{tab:main_results}
\begin{tabular}{lc}
\toprule
\textbf{Approach} & \textbf{Optimality gap} \\
\midrule
Generally trained PPO agent & $16.19$ \% \\
Pretrained DQN agent (PORL baseline) & $14.50$ \%\\
\ac{FDDMWR} & $13.63$ \% \\
Offline \ac{RL} agent & $12.48 \pm 0.63$ \% \\
Pretrained Offline \ac{RL} agent & $11.54 \pm 0.35$ \% \\
\bottomrule
\end{tabular}
\end{table}

To evaluate the robustness of the proposed pipeline against varying dataset quality, the agents are trained on three different datasets, each generated by one of the behavioral policies explained in section \ref{sec:behpolicies}. The results in \autoref{tab:dataset_ablation} show the effect of using heuristics, noisy-expert, and purely random datasets on both traditional offline \ac{RL} and our proposed \ac{PORL} agent. The results in \ref{fig:violinplot} show how dataset quality affects both approaches.

The proposed \ac{PORL} agent outperforms the standard offline CQL baseline across all dataset configurations. The performance gap increases as the quality of the behavioral policy decreases, particularly for the noisy-expert and fully random datasets.

\begin{table*}[tbp]
\centering
\caption{Makespan optimality gap (\%) for offline and PORL agents with different dataset compositions.}
\label{tab:dataset_ablation}
\begin{tabular}{lccc}
\toprule
\textbf{Approach} & \textbf{Combined Heuristics} & \textbf{Noisy-Expert} & \textbf{Random} \\
\midrule
Offline RL agent &$12.48 \pm 0.63$ \% & $16.89 \pm 2.15$ \%& $26.60 \pm 1.24$ \%\\
Pretrained Offline RL agent &$11.54 \pm 0.35$ \% & $13.19 \pm 0.88$ \% & $21.08 \pm 1.74$ \% \\
\bottomrule
\end{tabular}
\end{table*}

\section{Discussion and Limitations}

The results demonstrate that the proposed \ac{PORL} framework achieves lower optimality gaps than both the standalone offline \ac{RL} baseline and the considered online \ac{RL} approaches under substantial distributional shift. This suggests that knowledge acquired through online interaction with a simulator can be successfully transferred to an offline adaptation setting, despite differences between the source and target instance distributions.

A possible explanation for this improvement is that online pretraining exposes the agent to a broader range of scheduling states and actions than those contained in the offline dataset. As a result, the pretrained network provides more informative initial value estimates and state representations before offline optimization begins. In contrast, conventional offline \ac{RL} must learn these representations exclusively from the available dataset, making it more sensitive to limitations in dataset coverage and quality. The observed performance gains indicate that the pretrained policy acts as a useful prior that guides learning towards high-quality scheduling strategies while still allowing adaptation to the target distribution.

The results further show that the \ac{FDDMWR} dispatching rule outperforms both online \ac{RL} agents trained on uniformly distributed benchmark instances. This suggests that, in the evaluated setting, distributional shift may have a larger impact on performance than policy complexity. While the online \ac{RL} agents learn sophisticated scheduling strategies for the source distribution, these strategies do not fully transfer to the target shop floor. In contrast, dispatching rules are largely distribution-agnostic and therefore exhibit greater robustness under changing operating conditions. Nevertheless, both offline \ac{RL} and \ac{PORL} outperform the heuristic baseline, highlighting the benefit of adapting policies to production-specific data.

The reduced sensitivity to dataset quality observed in \ac{PORL} aligns with the broader offline \ac{RL} literature, which identifies dataset quality and coverage as a major influence on the performance \cite{levine_offline_2020,schweighofer_dataset_2022}. The results indicate that limiting divergence from the online pretrained policy contributes to maintaining PORL's superior performance with respect to Priority Dispatching Rules for 2 of the 3 datasets, while the performance from offline \ac{RL} degrades more rapidly as the dataset quality does.

Despite its superior performance, \ac{PORL} exhibits greater training instability than conventional offline RL. A likely explanation is the interaction between multiple optimization objectives. During fine-tuning, the DQN loss promotes value estimation, the CQL objective encourages conservative behavior with respect to the offline dataset, and the KL regularization term constrains deviations from the pretrained policy. When the pretrained policy and the behavioral policy underlying the dataset differ substantially, these objectives may provide conflicting optimization signals, leading to less stable training dynamics. In our experiments, this effect was partially mitigated through loss scheduling and hyperparameter ramping, although additional tuning complexity is introduced compared to standard offline RL.

While a full ablation study was outside the scope of this work, preliminary experiments without KL regularization resulted in catastrophic forgetting and substantially degraded performance, with optimality gaps reaching 25\%, indicating that the constraint plays an important role in preserving transferred knowledge.

The scope of the present evaluation is deliberately limited. First, the experiments are restricted to synthetic $10 \times 10$ \ac{JSSP} instances and only a single distribution-shift scenario. While the results indicate that \ac{PORL} can effectively adapt to specialized instance distributions, further validation on larger scheduling problems and real production datasets is required. Second, the proposed approach relies on the availability of a suitable simulation environment for pretraining. Significant mismatch between simulation and deployment conditions may reduce the effectiveness of the transferred policy. Finally, the impact of the KL regularization strength and alternative policy-constraining mechanisms remains an open research question and warrants further investigation.

\section{Conclusion}

This paper introduced \ac{PORL}, a hybrid training framework that combines simulation-based online \ac{RL} with offline adaptation to production-specific scheduling data. Unlike conventional hybrid \ac{RL} approaches that rely on offline pretraining followed by online fine-tuning, \ac{PORL} reverses this paradigm by first learning a general scheduling policy through online interaction and subsequently adapting it using offline \ac{RL}. To enable controlled adaptation, we further introduced a KL-divergence-based policy constraint that limits deviations from the pretrained policy during offline fine-tuning.

Experiments on \ac{JSSP} instances under substantial distributional shift demonstrated that \ac{PORL} consistently outperforms standalone offline RL, online \ac{RL} baselines, and priority dispatching rules. The results indicate that online pretraining provides informative prior knowledge that improves adaptation to specialized scheduling environments, reducing the dependence of offline RL on dataset quality and coverage.

These findings suggest that combining online exploration with offline adaptation is a promising direction for industrial scheduling applications where direct online learning is impractical, and standalone offline learning can be sensitive to the dataset quality and coverage. 

Future work will focus on evaluating \ac{PORL} on larger-scale scheduling problems and real-world production datasets, as well as investigating different levels of distribution shift and their impact on \ac{PORL} performance.

\bibliographystyle{ieeetr}
\bibliography{thesis}

@misc{nair_awac_2021,
	title = {{AWAC}: {Accelerating} {Online} {Reinforcement} {Learning} with {Offline} {Datasets}},
	shorttitle = {{AWAC}},
	url = {http://arxiv.org/abs/2006.09359},
	doi = {10.48550/arXiv.2006.09359},
	urldate = {2025-11-13},
	publisher = {arXiv},
	author = {Nair, Ashvin and Gupta, Abhishek and Dalal, Murtaza and Levine, Sergey},
	month = apr,
	year = {2021},
	note = {arXiv:2006.09359 },
}

@inproceedings{schweighofer_dataset_2022,
	title = {A {Dataset} {Perspective} on {Offline} {Reinforcement} {Learning}},
	issn = {2640-3498},
	url = {https://proceedings.mlr.press/v199/schweighofer22a.html},
	language = {en},
	urldate = {2025-11-19},
	booktitle = {Proceedings of {The} 1st {Conference} on {Lifelong} {Learning} {Agents}},
	publisher = {PMLR},
	author = {Schweighofer, Kajetan and Dinu, Marius-constantin and Radler, Andreas and Hofmarcher, Markus and Patil, Vihang Prakash and Bitto-nemling, Angela and Eghbal-zadeh, Hamid and Hochreiter, Sepp},
	month = nov,
	year = {2022},
	pages = {470--517},
}

@article{hajariwala_review_2025,
	title = {A {Review} of {Metaheuristic} {Algorithms} for {Job} {Shop} {Scheduling}},
	volume = {11},
	copyright = {Copyright (c) 2024},
	issn = {2730-4175},
	url = {https://ph02.tci-thaijo.org/index.php/mijet/article/view/254023},
	language = {en},
	number = {1},
	urldate = {2025-11-27},
	journal = {Engineering Access},
	author = {Hajariwala, Dharmik Chiragkumar and Patil, Srishti Sudhir and Patil, Sudhir Madhav},
	year = {2025},
	pages = {65--91},
}

@inproceedings{habib_zahmani_multiple_2015,
	title = {{Multiple} {Priority} {Dispatching} {Rules} {for} the {Job} {Shop} {Scheduling} {Problem}},
	doi = {10.1109/CEIT.2015.7232991},
	author = {Habib Zahmani, Mohamed and Atmani, Baghdad and Bekrar, Abdelghani and Aissani, Nassima},
    booktitle = {Proceedings of the 3rd International Conference on Control, Engineering and Information Technology},
	month = may,
	year = {2015},
}

@article{da_col_industrial-size_2022,
	title = {{Industrial}-{Size} {Job} {Shop} {Scheduling} with {Constraint} {Programming}},
	volume = {9},
	issn = {2214-7160},
	url = {https://www.sciencedirect.com/science/article/pii/S2214716022000215},
	doi = {10.1016/j.orp.2022.100249},
	urldate = {2025-12-04},
	journal = {Operations Research Perspectives},
	author = {Da Col, Giacomo and Teppan, Erich C.},
	month = jan,
	year = {2022},
	pages = {100249},
}

@article{brucker_branch_nodate,
	title = {A {Branch} and {Bound} {Algorithm} for the {Job}-{Shop} {Scheduling} {Problem}},
	language = {en},
	author = {Brucker, Peter and Jurisch, Bernd and Sievers, Bernd},
    journal = {Discrete Applied Mathematics},
    volume = {49},
    number = {1},
    pages = {107-127},
    year = {1994},
}

@article{artigues_branch_2008,
	title = {A {Branch} and {Bound} {Method} for the {Job}-{Shop} {Problem} with {Sequence}-{Dependent} {Setup} {Times}},
	volume = {159},
	issn = {1572-9338},
	url = {https://doi.org/10.1007/s10479-007-0283-0},
	doi = {10.1007/s10479-007-0283-0},
	language = {en},
	number = {1},
	urldate = {2025-12-12},
	journal = {Annals of Operations Research},
	author = {Artigues, Christian and Feillet, Dominique},
	month = mar,
	year = {2008},
	pages = {135--159},
}

@misc{levine_offline_2020,
	title = {Offline {Reinforcement} {Learning}: {Tutorial}, {Review}, and {Perspectives} on {Open} {Problems}},
	shorttitle = {Offline {Reinforcement} {Learning}},
	url = {http://arxiv.org/abs/2005.01643},
	doi = {10.48550/arXiv.2005.01643},
	urldate = {2025-12-12},
	publisher = {arXiv},
	author = {Levine, Sergey and Kumar, Aviral and Tucker, George and Fu, Justin},
	month = nov,
	year = {2020},
	note = {arXiv:2005.01643 },
}

@inproceedings{nababan_branch_2008,
	title = {{Branch} and {Bound} {Algorithm} in {Optimizing} {Job} {Shop} {Scheduling} {Problems}},
	volume = {1},
	issn = {2155-899X},
	url = {https://ieeexplore.ieee.org/document/4631564/},
	doi = {10.1109/ITSIM.2008.4631564},
	urldate = {2026-01-31},
	booktitle = {2008 {International} {Symposium} on {Information} {Technology}},
	author = {Nababan, Erna Budhiarti and Hamdan, Abdul Razak and Abdullah, Salwani and Zakaria, Mohamad Shanudin},
	month = aug,
	year = {2008},
	pages = {1--5},
}

@misc{munikoti_challenges_2022,
	title = {Challenges and {Opportunities} in {Deep} {Reinforcement} {Learning} with {Graph} {Neural} {Networks}: {A} {Comprehensive} {Review} of {Algorithms} and {Applications}},
	shorttitle = {Challenges and {Opportunities} in {Deep} {Reinforcement} {Learning} with {Graph} {Neural} {Networks}},
	url = {http://arxiv.org/abs/2206.07922},
	doi = {10.48550/arXiv.2206.07922},
	urldate = {2026-02-05},
	publisher = {arXiv},
	author = {Munikoti, Sai and Agarwal, Deepesh and Das, Laya and Halappanavar, Mahantesh and Natarajan, Balasubramaniam},
	month = nov,
	year = {2022},
	note = {arXiv:2206.07922 },
}

@inproceedings{maharjan_reinforcement_2026,
	title = {A {Reinforcement} {Learning} {Environment} for {Job} {Shop} {Scheduling} with {Tool} {Management}},
	isbn = {978-3-032-11442-6},
	doi = {10.1007/978-3-032-11442-6_6},
	language = {en},
	booktitle = {Artificial {Intelligence} {XLII}},
	publisher = {Springer Nature Switzerland},
	author = {Maharjan, Reshma and Andersen, Per-Arne and Jiao, Lei},
	editor = {Bramer, Max and Stahl, Frederic},
	year = {2026},
	pages = {80--93},
}

@article{park_learning_2021,
	title = {{Learning} to {Schedule} {Job}-{Shop} {Problems}: {Representation} and {Policy} {Learning} {Using} {Graph} {Neural} {Networks} and {Reinforcement} {Learning}},
	volume = {59},
	issn = {0020-7543, 1366-588X},
	shorttitle = {Learning to schedule job-shop problems},
	url = {http://arxiv.org/abs/2106.01086},
	doi = {10.1080/00207543.2020.1870013},
	number = {11},
	urldate = {2026-02-05},
	journal = {International Journal of Production Research},
	author = {Park, Junyoung and Chun, Jaehyeong and Kim, Sang Hun and Kim, Youngkook and Park, Jinkyoo},
	month = jun,
	year = {2021},
	note = {arXiv:2106.01086 },
	pages = {3360--3377},
}

@article{remmerden_offline_2025,
	title = {Offline {Reinforcement} {Learning} for {Learning} to {Dispatch} for {Job} {Shop} {Scheduling}},
	volume = {114},
	issn = {1573-0565},
	url = {https://doi.org/10.1007/s10994-025-06826-w},
	doi = {10.1007/s10994-025-06826-w},
	language = {en},
	number = {8},
	urldate = {2026-02-27},
	journal = {Machine Learning},
	author = {Remmerden, Jesse van and Bukhsh, Zaharah and Zhang, Yingqian},
	month = jul,
	year = {2025},
}

@article{echeverria_offline_2025,
	title = {Offline {Reinforcement} {Learning} for {Job}-{Shop} {Scheduling} {Problems}},
	volume = {184},
	issn = {1568-4946},
	url = {https://www.sciencedirect.com/science/article/pii/S156849462501049X},
	doi = {10.1016/j.asoc.2025.113736},
	urldate = {2026-02-27},
	journal = {Applied Soft Computing},
	author = {Echeverria, Imanol and Murua, Maialen and Santana, Roberto},
	month = dec,
	year = {2025},
	pages = {113736},
}

@misc{fujimoto_off-policy_2019,
	title = {Off-{Policy} {Deep} {Reinforcement} {Learning} without {Exploration}},
	url = {http://arxiv.org/abs/1812.02900},
	doi = {10.48550/arXiv.1812.02900},
	urldate = {2026-03-05},
	publisher = {arXiv},
	author = {Fujimoto, Scott and Meger, David and Precup, Doina},
	month = aug,
	year = {2019},
	note = {arXiv:1812.02900 },
}

@misc{kumar_conservative_2020,
	title = {Conservative {Q}-{Learning} for {Offline} {Reinforcement} {Learning}},
	url = {http://arxiv.org/abs/2006.04779},
	doi = {10.48550/arXiv.2006.04779},
	urldate = {2026-03-05},
	publisher = {arXiv},
	author = {Kumar, Aviral and Zhou, Aurick and Tucker, George and Levine, Sergey},
	month = aug,
	year = {2020},
	note = {arXiv:2006.04779 },
}

@misc{remmerden_generalizing_2025,
	title = {Generalizing {Beyond} {Suboptimality}: {Offline} {Reinforcement} {Learning} {Learns} {Effective} {Scheduling} through {Random} {Data}},
	shorttitle = {Generalizing {Beyond} {Suboptimality}},
	url = {http://arxiv.org/abs/2509.10303},
	doi = {10.48550/arXiv.2509.10303},
	urldate = {2026-03-10},
	publisher = {arXiv},
	author = {Remmerden, Jesse van and Bukhsh, Zaharah and Zhang, Yingqian},
	month = sep,
	year = {2025},
	note = {arXiv:2509.10303 },
}

@article{chen_offline--online_2025,
	title = {An {Offline}-to-{Online} {Reinforcement} {Learning} {Framework} with {Trajectory}-{Guided} {Exploration} for {Industrial} {Process} {Control}},
	volume = {154},
	issn = {0959-1524},
	url = {https://www.sciencedirect.com/science/article/pii/S0959152425001635},
	doi = {10.1016/j.jprocont.2025.103535},
	urldate = {2026-03-12},
	journal = {Journal of Process Control},
	author = {Chen, Jiyang and Luo, Na},
	month = oct,
	year = {2025},
	pages = {103535},
}

@misc{zhang_learning_2020,
	title = {Learning to {Dispatch} for {Job} {Shop} {Scheduling} via {Deep} {Reinforcement} {Learning}},
	url = {http://arxiv.org/abs/2010.12367},
	doi = {10.48550/arXiv.2010.12367},
	urldate = {2026-03-17},
	publisher = {arXiv},
	author = {Zhang, Cong and Song, Wen and Cao, Zhiguang and Zhang, Jie and Tan, Puay Siew and Xu, Chi},
	month = oct,
	year = {2020},
	note = {arXiv:2010.12367 },
}

@article{yao_control_2023,
	title = {Control of {Hybrid} {Electric} {Vehicle} {Powertrain} {Using} {Offline}-{Online} {Hybrid} {Reinforcement} {Learning}},
	volume = {16},
	copyright = {http://creativecommons.org/licenses/by/3.0/},
	issn = {1996-1073},
	url = {https://www.mdpi.com/1996-1073/16/2/652},
	doi = {10.3390/en16020652},
	language = {en},
	number = {2},
	urldate = {2026-03-30},
	journal = {Energies},
	publisher = {Multidisciplinary Digital Publishing Institute},
	author = {Yao, Zhengyu and Yoon, Hwan-Sik and Hong, Yang-Ki},
	month = jan,
	year = {2023},
	pages = {652},
}

@misc{xu_how_2019,
	title = {How {Powerful} are {Graph} {Neural} {Networks}?},
	url = {http://arxiv.org/abs/1810.00826},
	doi = {10.48550/arXiv.1810.00826},
	urldate = {2026-06-10},
	publisher = {arXiv},
	author = {Xu, Keyulu and Hu, Weihua and Leskovec, Jure and Jegelka, Stefanie},
	month = feb,
	year = {2019},
	note = {arXiv:1810.00826},
}

@misc{hoss_scalable_2026,
	title = {Scalable {Production} {Scheduling}: {Linear} {Complexity} via {Unified} {Homogeneous} {Graphs}},
	shorttitle = {Scalable {Production} {Scheduling}},
	url = {http://arxiv.org/abs/2604.23841},
	doi = {10.48550/arXiv.2604.23841},
	urldate = {2026-06-10},
	publisher = {arXiv},
	author = {Hoss, Jonathan and Link, Moritz and Klarmann, Noah},
	month = apr,
	year = {2026},
	note = {arXiv:2604.23841},
}
\end{document}